# MLOps: A Review

Samar Wazir[1], Gautam Siddharth Kashyap[2], Parag Saxena[3]

[1]samar.wazir786@gmail.com, [2]officialgautamgsk.gsk@gmail.com, [3]psaxena4@uncc.edu

[1,2]Department of Computer Science and Engineering, SEST, Jamia Hamdard, New Delhi, India
[3]School of Data Science, University of North Carolina Charlotte, USA

*Abstract*—Recently, Machine Learning (ML) has become a widely accepted method for significant progress that is rapidly evolving. Since it employs computational methods to teach machines and produce acceptable answers. The significance of the Machine Learning Operations (MLOps) methods, which can provide acceptable answers for such problems, is examined in this study. To assist in the creation of software that is simple to use, the authors research MLOps methods. To choose the best tool structure for certain projects, the authors also assess the features and operability of various MLOps methods. A total of 22 papers were assessed that attempted to apply the MLOps idea. Finally, the authors admit the scarcity of fully effective MLOps methods based on which advancements can self-regulate by limiting human engagement.

## I. INTRODUCTION

Recent improvements in big data, ML and AI have been widely applied in a variety of areas [1], [2]. ML models are commonly used in areas that help with speech or face recognition. However, developing and applying ML methods is more difficult and complex than using traditional methods for the evaluation and verification of a model. Therefore, it may return to an earlier stage of development, altering the model's performance. As numerous stages are involved in developing ML methods that need varying levels of labour. For example, the development stage includes model development, design, assessment, feature extraction, data validation and processing, etc [3]. To manage these stages, the following things are required:

- Significant data centres are required in the model development stage in the form of a Graphics Processing Unit (GPU), Tensor Processing Unit (TPU), and Central Processing Unit (CPU).
- Sometimes a model failure occurs as a result of a variety of factors, including changes in the data distribution settings, and changes in the model retraining mechanism.

In addition to the above, massive databases for model testing and training provide data management and storage challenges [4], [5]. Therefore, to solve these challenges posed by computational models, there is a need for rapid and dependable ML methodologies, as well as infrastructure setup. ModelOps, Kubeflow, and TensorFlow Extended (TFX) [4], [6], [7] are examples of systems that provide end-to-end development control for ML methods by arranging their stages into different ML pipelines. Several significant mechanisms and services are provided to handle pipeline installation, data storage, task planning, and work control. Software Development Kit (SDK) or Kubeflow's and ModelOps pipeline structure allows developers to set ML performance levels. These pipeline libraries may comprise steps that correspond to the stages, such as data investigation, model building, setup, and assessment. ML pipelines leverage Development and Operations (DevOps) methods to not only performance levels but also automates them. The advantages of Continuous Delivery (CD) and Continuous Integration (CI), can improve engine efficiency and faster code transfer, and automate the growth and setup of ML methods.

MLOps is a word that describes the fusion of ML operation (Ops) and development (Dev) created by applying DevOps codes to ML methods. These structures and stages have provided well-designed modules and functions, as well as ML programmes, to help reduce the ongoing maintenance costs associated with ML systems [8]. Because of virtualization and the cloud, the development control of ML systems benefits from ML pipeline platforms in terms of productivity and dependability. However, little is known about how these systems work, such as how much computing power they require or how long it takes to construct a model. Therefore, this study aims to contribute to and respond to some of the issues raised above. The following is the paper's key contribution:

1. This study aims to provide a background on DevOps and MLOps, as well as the ML lifecycle, to individuals who are unfamiliar with these terminologies.
2. A thorough examination of 22 papers that employed the MLOps.
3. The authors also compare and assess the features and operability of MLOps tools to determine the most suited tool structure for certain projects.
4. Finally, this study discusses the MLOps' future directions.

The motivation for writing this paper will be discussed in the next section. The background of DevOps and MLOps is discussed in section III. The ML lifecycle is described in section IV of the paper. The literature review of studies is described in section V. The MLOps tools stack is discussed in section VI. In section VII, a comparison of different MLOps tool stacks is made. The author's opinions are offered in the discussion portion of section VIII, and the paper is concluded with some conclusions in section IX.

## II. MOTIVATIONS

With the rise in popularity of CD and CI [9], [10], and, in particular, DevOps [11], it's becoming more common to build up ML systems during the actual development period. MLOps advocates for inspection and automation at all stages of the development and implementation of ML methods, including combination, experimentation, importing, set-up, and substructure control. Consider the procedures necessary to design and set up ML systems to grasp the MLOps issues [12] such as in the beginning, data must be available for development. For partitioning data, there are a few well-known approaches such as: 1) Training, 2) Testing, and 3) Cross-validation. An ML model must then be chosen, together with its hyperparameters. The model is then trained using the training data. Throughout the training stage, the model is tweaked repeatedly until the outcome matches the "right answers" in the development. To validate this competent model, various types of data might be used. If the authentication is successful (based on whatever criteria we set), the model, like any other module, is ready to be installed. After they've been installed, ML systems, like any other system, need to be monitored. However, ML systems must account for errors and ideas that may evolve. In other respects, the model might continue to improve even as it is being used. As a result, the ML systems must be tailored to these specific needs. To summarise, the ML phases that occur before the completion of the ML model appear to be waterfall-like, however, the methods associated with traditional software are used to implement the model on a larger entire chart.

## III. BACKGROUND OF DEVOPS AND MLOPS

In this section, we'll go over the background of DevOps and MLOps, which includes the following:

### A. Overview of DevOps

The DevOps-based software method aids in increasing the supply stage and occurrence rate while preserving quality, stability, and privacy [13]. Many businesses are automating their supply chains while preserving them as a link between software development and software delivery, bringing together the growth and process divisions. As a result, DevOps is a never-ending process including constant growth, combination, distribution, and verification [14]. DevOps is a set of methods for detecting changes in software tools and managing the reliability of other tools affected by the conversion to sustain solution supply productivity and speed [15], [16]. Many Integrated Development Environments (IDEs) are used during the software development process. For example, Git[1] for style control and a Docker[2] package for packaging all needs and libraries, and Jenkins[3], for permitting the combining of programmes contributed by a large number of operators. As a result, Continuous Integration and Continuous Delivery (CICD) employs a range of strategies to ensure that software releases are quick and reliable [17]. Changes in client needs, social circumstances, and technical issues are all common causes for software entities to evolve. The automation technique must be able to detect these changes, assess their impact, and maintain reliability. Because the DevOps method is integrated, therefore, those technologies should improve network skills and visualization [18]. The availability of automated techniques and technical skills helps to reduce the amount of work required throughout the software development cycle [19], resulting in the elimination of unnecessary costs.

### B. Overview of MLOps

The MLOps method incorporates ML models into the software development process. As shown in Fig. 1, it integrates ML features with DevOps concepts, allowing the automatic installation and effective monitoring of ML models in the development setting. MLOps systems should be capable of working as a collective, constant, repeatable, validated, and monitored in a way to meet corporate MLOps objectives. The three main components of the MLOps development process are model, code, and data. The MLOps structures[4] require technique automation to keep the development phase running. As a result, MLOps makes software development more accessible and faster while posing less risk. Productivity is aided by quick model building, high-quality ML models, and quick placement and production. The MLOps methods allow us to manage and screen a large number of models thanks to CICD's extensibility. The high level of collaboration between teams helps to avoid conflicts and expedite the transfer procedure. Controlling identifiability and reliability also helps to reduce the hazards that come with it.

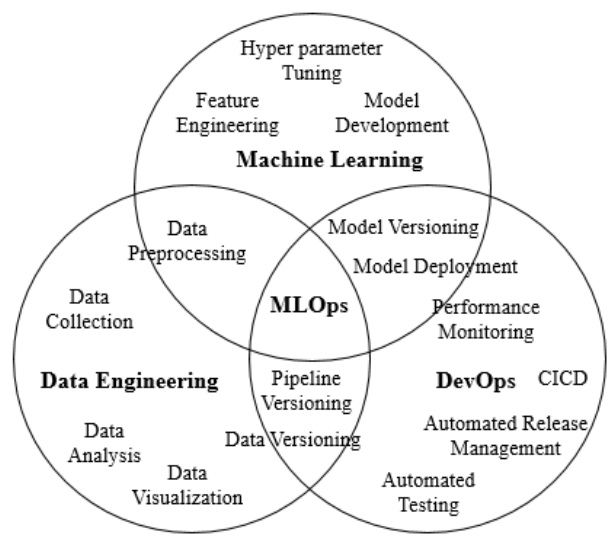

Fig. 1. Overview of MLOps

---

[1] https://www.atlassian.com/git/tutorials/comparing-workflows/gitflow-workflow
[2] https://www.docker.com/resources/what-container/
[3] https://plugins.jenkins.io/uno-choice/
[4] https://ml-ops.org/

## IV. MACHINE LEARNING LIFECYCLE

The model enhancement methodology comprises a system i.e. CICD, a tool for integrating ML into the software development process. These training models are related to several factors, including a set of rules and hyperparameters that are algorithmically changed. After installation, the model should be checked regularly to ensure that any productivity loss is avoided. Several characteristics, specialist groups, and approaches are used during the development phase. The ML lifecycle phase includes processes such as model requirements, data gathering, and early creation, as well as growth assessment, and one-to-one testing.

Data is a factor that influences the overall performance of an ML model [20]. Data might be private or public, and it can be gathered by tests or studies. Because of mistakes and data excess, data should be filtered and initialised before being used for training [21]. After that, feature engineering methodologies are utilised to locate and detect crucial data properties to create ML models [22]. Hyper-parameter adjustment and the most successful strategies are implemented before the training operation. A source is preserved to manage the scripts and models. The model build phase is started using DevOps concepts once the model is devoted to the source. The integration and unit testing phases will be utilised in combination with the model build phase. Validation and analysis of the model are also required to determine the model's function. After the model has passed the accuracy phase, it is deployed in the development environment. Correct one-to-one testing should be supported in the same way as traditional software development. Fig. 2 shows the level vice interface perspective of the ML, DevOps, and information flow from the MLOps practice. MLOps is thus defined as a method for automating the ML lifecycle phase by removing human participation from repetitive operations. Typically, the MLOps cycle begins with the business examination team and expertise asking commercial inquiries and researching requirements. Engineers select the type of model to be created, the functions that must be monitored, and the data collection system and its accessibility based on the requirements. As a result, various jobs such as software and data experts, as well as engineers, are employed in each of these stages to meet the goals as outlined in Fig. 3. Data and software engineers, as well as feature declaration experts, concentrate on the unavoidable variables that arise during the execution of a release before it is sent to creation. Data experts use DevOps ideas and technologies to expand development while taking into account scalability, security, and reliability metrics. Finally, DevOps methodologies are employed to conduct continuous one-on-one testing and evaluation.

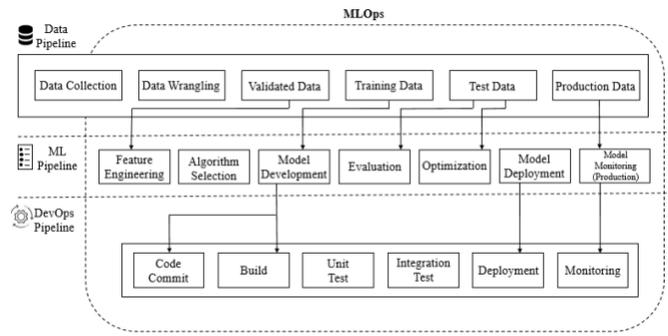

Fig. 2. MLOps flow

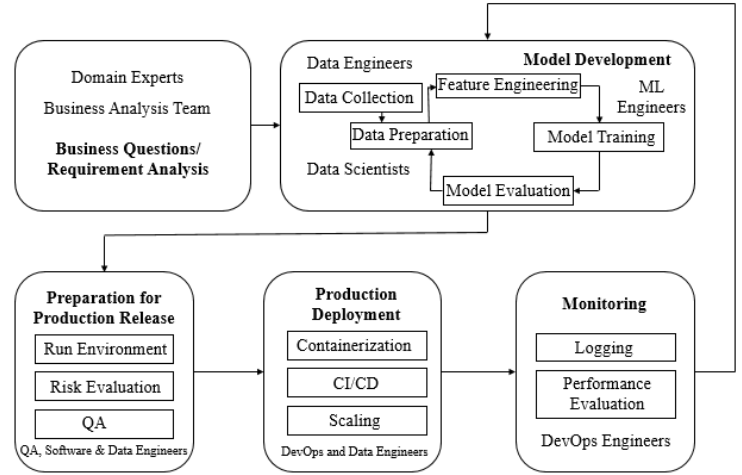

Fig. 3. MLOps phases

## V. LITERATURE REVIEW

Several studies have been published such as Fursin [23] offered an overview of the Collective Knowledge (CK) platform leveraging MLOps, reusable best practices, and open Application Programming Interfaces (APIs). In contrast, CK is still a proof-of-concept with a large number of upgrade stages. He will work on research methodologies in the future to make it easier to deploy, such as standardising JavaScript Object Notation (JSON) and API meta reports for all CK modules and frameworks. Simultaneously, the desired outcome was reached with the best transaction in terms of energy, latency, accuracy, speed, and size. Moreschini et al. [24] developed MLOps for evolvable AI-intensive software systems. The goal is to create a self-maintaining ML-based constructed module that can improve in unison with software advancements. Their long-term goal is to have a better understanding of how MLOps work in the commercial world. Renggli et al. [25] proposed an MLOps based on data quality. Their main goal is to show how different data quality attributes are spread across ML lifecycle phases. In terms of models, they faced several difficulties. They propose two future approaches that, in their opinion, are required to make MLOps a necessary competency for the primary exploration field. Granlund et al. [26] offered a multi-organization configuration in two real-world scenarios using MLOps challenges. The highlighted limits are related to scale and combined, with real-life scenarios demonstrating how ML has been applied. They will also collaborate on a joint operating strategy involving all of the official doms. Mäkinen et al. [27] investigated the current state-of-the-art

ML to determine how much MLOps is necessary for today's ML systems and how much software engineers are still dealing with numbers and demonstrating issues without considering implementation. They plan to learn more about MLOps in this scenario in the future. Min et al. [28] proposed SensiX++, a multi-tenant runtime for the responsive model implemented on edge devices such as Internet of Things (IoT) sensors, microphones, and cameras, with integrated MLOps. Their experiments on an edge device demonstrated how simple it is to design and coordinate various sight and sound models, as well as the benefits of the machine supplied by the device's core components. Van Der Goes [29] proposes an MLOps method in addition to two related functions. He demonstrated how to develop a self-service ML workflow and a recommender system that can grow to new branches faster and with less technological debt. He will also improve the methods that are currently being used to achieve optimum results. Borg [30] proposed utilising MLOps and notebook interfaces with rebars and buttresses to achieve agility in software 2.0. He started by talking about notebook interfaces, their shortcomings, and a proposal to reduce the standards so that they can coexist with current IDEs. Second, he talked about his opinions on MLOps and his present work in this situation to deliver AI-based strengthened development. MLOps definitions, tools, and challenges were identified by Symeonidis et al. [31]. The goal of their research is to gain insight into the subject of MLOps. The major purpose of their research is to sketch out the workings and characteristics of such technologies while highlighting current difficulties and developments. They want to conduct a more comprehensive and practical examination of the use of MLOps with AutoML in the future. Hewage and Meedeniya [32] used tool support to implement the MLOps survey. They looked into the existing technical issues with software design and delivery in companies working on ML. They also acknowledge that a fully effective MLOps framework capable of automating activities while reducing human involvement is few. Model and dataset registries can be monitored and controlled, allowing them to be reused and modified for future databases. Operational obstacles for widespread edge AI deployment using MLOps and TinyMLOps are studied by Leroux et al. [33]. They focused on the technical side of things and discovered several issues that the distributed edge setup either causes or exacerbates. They plan to work on more advanced frameworks and technologies in the future. MLOps' definition, design, and overview were suggested by Kreuzberger et al. [34]. The primary goal of this research is to provide insight into the MLOps and its tools. They also gave an overview of MLOps and a list of open issues in the field. Finally, they make suggestions for ML scientists and specialists who want to automate and run their ML tools using a specific set of technologies. Klein et al. [35] proposed MLOps, DataOps, and SageOps for data through DevOps faster-time-to-knowledge. This research also looks at the standard framework's technical implementation for design and deployment solutions, as well as the standard framework's operations and outcomes, as well as future research and development aims. Ruf et al. [36] created a method for selecting open-source tools while demystifying MLOps. This research demonstrated MLOps' potential. With the introduction of numerous characters and roles engaged in MLOps processes, an outline of duties and support tools for the various stages was presented. Future research might look into the potential of AutoML features for MLOps procedures, as well as how they're connected and managed inside these workflows. Melgar et al. [37] made use of Ease.ML for MLOps and MLDev lifecycle management system. The most recent version of Ease.ML features i.e. an eight-step method that covers the entire software architecture inspired by the challenges and misunderstandings found during user engagement. These flaws, they believe, indicate intriguing future study directions. Gürses-Tran and Monti [38] offered new advancements in time series forecasting for MLOps in power systems. When adjusting the quality of a certain model, prediction makers primarily consider error terms and failure metrics. Temporal fusion transformer models are discussed in this work as intriguing choices for future research. Raj et al. [39] developed an automation framework for IoT applications using edge MLOps. They presented a complete development cycle for automating IoT activities by coordinating two key pipelines: interface inference and cloud management, to achieve adaptability. They will investigate the performance of their framework in cases where the AI setup is more time-sensitive, such as a brief training phase, in the future. To improve the short-term energy consumption forecast, Fujii et al. [40] used a digital twin structural design combined with MLOps methodologies. However, their next research focus will be on how ontologies for home automation promote interactive dialogues and provide personalised recommendations through user testing. For Successive Interference Cancellation (SIC)-based least setting length with better power development for Underwater Acoustic Networks (UANs), Li et al. [41] used MLOps. The proposed MLOps technique considers two types of traffic loads, hence it is separated into two sub-algorithms: Weighted Traffic Load (WMLOPS) and Unified Traffic Load (UMLOPS). In the future, they will look into an optimal polynomial solution for K-SIC. For vision-based inspection development in manufacturing, Lim et al. [42] built an MLOps lifecycle structure. They identified four key users and five aspects to complete the MLOps development cycle. The system's ML models outperform traditional inspection methods, with the purpose of modernising process automation in future work. Banerjee et al. [43] designed an MLOps pipeline to handle the five challenges in their accomplishment monitoring use scenario. They also give and aim for a fully automated method to generate new models on a continuous basis using the most up-to-date data. They also use some new technology techniques for this research. By implementing a DevOps experience with an ML platform on a common real object configuration, Zhou et al. [44] demonstrated the practicality of constructing operational ML pipelines utilising established CD/CI tools and ML platforms like Kubeflow, which can continually reskill models when certain scenarios emerge. They looked at the cost and labour use of each stage of the ML pipeline, as well as the use of the ML platform and analytical models, and suggested identifying potential roadblocks, such as GPU utilisation. Their work can be used as a reference for building ML pipeline platforms in the real world.

## VI. MLOps Tools Stack

The MLOps application stack simplifies, improves, and accelerates the ML lifecycle process.

### A. Kubeflow

Kubeflow[5] is a Google project focused on deploying ML models to manage simple, adaptive, and efficient instals that meet the requirements. It is an open-source ML platform that structures the pieces of the ML system on top of the Kubernetes platform and provides good efficiency, installation, and monitoring of ML applications throughout their lifecycle using standardised pipelines. Several ML methods and monitoring tools are supported by Kubeflow such as notebook servers, training operators, and KFServing are all included, as well as a communicative User Interface (UI). Despite the lack of a dedicated solution for the CICD workflow, Kubeflow pipelines can be used to establish repetitive work strategies that automate the procedures required to develop an ML process, ensuring stability, assisting with debugging and compliance requirements, and reducing loop time[6].

### B. MLflow

MLflow[7] is a non-cloud, open-source platform for managing the end-to-end ML development process, which includes four core tasks:

1. **Monitoring**: By clicking logs and queries about all the contributions, management, and results, MLflow allows operators to monitor trials to capture and relate factors and outcomes.
2. **Development plans:** The MLflow development plans can be used as a container for ML code, making it easier to reuse and duplicate [45].
3. **Frameworks:** MLflow frameworks allow us to handle a high number of ML libraries, model interference and allocation platforms.
4. **Model record:** The MLflow model record portion assists the critical method supplies in collaboratively managing the entire development process of an ML model, including management, segment deviations, and remarks.

MLflow is also capable of executing and controlling any ML programming language and library. Additionally, using Azure ML, Apache Spark 11, and Amazon Web Service (AWS) SageMaker, it is feasible to help models and set up machine-to-machine communication, achieving CICD goals using cloud service features. It also enables statistical analysis of deployed models [43]. The lack of implicit books and the inability to maintain book form when used as an IDE for enhancement are two of this tool's shortcomings. MLFlow also does not keep user organisation and does not provide full updation, unlike clustering tests [36].

### C. Iterative Enterprise

The components of the iterative enterprise[8] that handle and operate ML databases, tests, and models include Data Version Control (DVC) and Continuous Machine Learning (CML). When it comes to MLOps, DVC is critical, but it can be tough to maintain when the database is large. DVC is a platform-agnostic and open-source version system for ML tools that allows users to develop shared, uniform ML models while keeping track of model, pipeline, and record-type behaviour. It can also build tiny metafiles to handle and achieve enormous files, measurements, models, databases, applications, and research findings to make the most of versioning. CML supports CICD for ML projects. It uses GitHub or GitLab to arrange ML tests, auto-generate data, and keep track of changes using graphs and metrics in each Git pull application. DVC studio also facilitates effective team collaboration and information sharing.

### D. DataRobot

The DataRobot's[9] MLOps platform provides a centralised development location where models in development may be put up, screened, and handled regardless of how, where, or when they were established. It contains a well-organized registry where all development models may be stored and managed. DataRobot covers the ML development phases from growth to use. It also helps with a variety of libraries, programming languages, and progress settings by managing code sources. Single operators, on the other hand, are responsible for purchasing permits for each sample of the defined convention.

### E. Allegro.ai (Clear ML)

For quick product delivery, Allegro.ai offers open-source MLOps tools. ClearML[10] from Allegro.ai is a separate slot for exploring, coordinating, setting up, and constructing data stores. ClearML's major steps are: test, DataOps, coordinate, remote, hyper-datasets, and set up. ClearML also comes with several modules. The codebase and structure are combined in the ClearML python package, for example. ClearML Server is a server that records tests, workflows, and data models while managing MLOps. The ClearML manager provides rates for adaption and repetition. You can use the ClearML session module to execute VSCode and Jupyter Notebooks from a remote location.

### F. MLReef

---

[5] https://www.kubeflow.org/
[6] https://ml-ops.org/content/state-of-mlops
[7] https://mlflow.org/
[8] https://iterative.ai/
[9] https://www.datarobot.com/
[10] https://clear.ml/

MLOps is a git-based open-source project. This platform includes MLReef[11], which provides a centralised location for managing the ML development process. This software manages work in repositories so that ML can be improved in a repeatable, efficient, and collaborative manner. Because of its rapid and furious capabilities, cooperation, free CPU/GPU obtainability, allotment, and repeatability, MLReef is preferred among MLOps platforms.

*G. Streamlit*

Streamlit[12] is a Python tool that accelerates web app development. It has a straightforward user interface that does not necessitate any backend configuration. We can iterate on our code and see the results as we go with Streamlit. Operators can utilise the Streamlit cloud tools to deploy the added web server and visualise the web app's performance. Because it is a python module that provides an increased impact for data visualisation, it may be used for dashboard progress projects in general.

*H. MLOps with Could Service Providers*

MLOps, unlike DevOps, has no proven results, thus it uses a set of applications to automate the process and still requires human participation frequently. In general, cloud service providers provide ML platforms, such as Google Cloud's AI Platform, AzureML studio, and Amazon SageMaker, to help ML solutions to become more productive. They also offer ways for customers who aren't familiar with AI to quickly learn how to use ML. Consumers are also encouraged to join cloud service providers' ML platforms because of the Pay-As-You-Go pricing model. To aid MLOps, Microsoft Azure provides the following modules[13].

- Azure ML contains created Notebooks and the ability to generate, enhance, and check a wide range of models regularly, regardless of skill level.
- Azure Monitor collects and analyses metrics to help us operate more efficiently.
- Azure Kubernetes Service.
- Azure Pipelines standardises ML pipelines for creating and testing applications.

Building the MLOps location on the Google cloud service provides the following features[14]:

- **Dataflow:** A data-management service that retrieves, modifies, and generates models based on records.
- **AI notebook:** An environment for generating models that includes a programming environment.
- **TFX:** Skilled in installing ML pipelines.
- **Kubeflow pipelines:** Systematizing ML setting out on top of Google Kubernetes Engine.

---

[11] https://about.mlreef.com/
[12] https://streamlit.io/
[13] https://azure.microsoft.com/en-us/services/machine-learning/mlops/
[14] https://cloud.google.com/architecture/setting-up-an-mlops-environment

- **Cloud Develop:** Form, experiment, and install apps.

MLOps on AWS can be implemented using Amazon SageMaker, a fully operational structure that can handle the ML development process by systematising MLOps methods. It allows for more effective and dynamic training, handling, construction, use, and testing of ML applications.

## VII. COMPARISON OF MLOPS TOOLS

Through study, new frontiers in the creation of advanced MLOps systems have been explored. Although there are several solutions to help with structural facts identifiability DevOps training [46], there are no essential technologies that manage identifiability in the MLOps development process. Several studies [47], [48] have demonstrated how to employ automation technologies to maintain structural facts consistency in DevOps-based project management. This technological concept could also be used to maintain MLOps' identifiability. Because of the challenges and time-consuming manual development that must be adjusted regularly, only a few research have looked into the reliability of MLOps [49]. The establishment of MLOps technology structures has been the focus of the majority of extant research [50]. Some low-cost solutions, like Kubeflow and MLflow, can provide similar functionalities with automated operations to some extent. These tools will also make the growth process go more smoothly and assist us to estimate the quantity of work needed. Table I illustrates a comparison of the characteristics of the current MLOps platform. The functions of Hyperparameter Tuning (HT), Data Versioning (DV), Pipeline Versioning (PV), Model Deployment (MD), CICD availability, Performance Monitoring (PM), and Model and Experiment Versioning (MEV) were used to compare the MLOps systems. This might be mentioned as well when selecting a platform for clarifying progress settings.

Table I. Feature comparison of existing platforms

|  | HT | DV | PV | MD | CICD | PM | MEV |
|---|---|---|---|---|---|---|---|
| Kubeflow | ✔ |  |  | ✔ | ✔ | ✔ | ✔ |
| MLflow | ✔ | ✔ | ✔ | ✔ | ✔ |  | ✔ |
| Iterative Enterprise |  | ✔ |  | ✔ | ✔ | ✔ | ✔ |
| DataRobot | ✔ |  |  | ✔ |  | ✔ | ✔ |
| ClearML |  | ✔ | ✔ | ✔ | ✔ | ✔ | ✔ |
| MLReef | ✔ | ✔ | ✔ | ✔ | ✔ | ✔ | ✔ |
| Streamlit |  | ✔ |  | ✔ |  | ✔ | ✔ |
| AWS SageMaker | ✔ | ✔ | ✔ | ✔ | ✔ | ✔ | ✔ |

In addition, different software development environments use a variety of libraries, frameworks, and programming languages. As a result, an MLOps system should be able to provide platform-independent services. In this case, structures, libraries, and supported languages should all be considered when choosing an appropriate MLOps system. Table II lists the languages supported by various MLOps systems. As a result, while AWS SageMaker and MLflow beat the competition, they also have faults that need to be addressed, as described under each of the structures above. Although cloud service providers provide similar solutions, they are more expensive and do not directly address the ML difficulty through a dashboard. In addition, certain platforms do not offer free licences for use as embedded structures.

Table II. Language-support comparison of existing platforms

|  | PyTorch | Jupyter Notebook | Java | TensorFlow | Scikit-Learn | Keras | R | Python |
|---|---|---|---|---|---|---|---|---|
| Kubeflow | ✔ | ✔ |  | ✔ | ✔ |  |  | ✔ |
| MLflow | ✔ |  | ✔ | ✔ | ✔ | ✔ | ✔ | ✔ |
| Iterative Enterprise |  |  |  | ✔ | ✔ |  |  | ✔ |
| DataRobot | ✔ |  | ✔ | ✔ | ✔ | ✔ | ✔ | ✔ |
| ClearML | ✔ | ✔ |  | ✔ | ✔ | ✔ |  | ✔ |
| MLReef | ✔ |  |  | ✔ | ✔ | ✔ |  | ✔ |
| Streamlit | ✔ |  |  | ✔ |  | ✔ |  | ✔ |
| AWS SageMaker | ✔ | ✔ | ✔ | ✔ | ✔ | ✔ | ✔ | ✔ |

## VIII. DISCUSSION

The format of data, driving method, evaluation matrices, training level, and augmentation are all factors that influence the accuracy of ML method predictions. Precipitation models, for example, need the most up-to-date real-time data and are constantly retrained to provide more precise forecasts. As a result of using reproducible workflows, developing datasets should be restructured without requiring human intervention. It's tough to use MLOps to systematise these supervisory procedures. In addition, the MLOps platform should be capable of generating frameworks and processes, as well as allowing them to be easily reused and copied to expand the scope of testing and get the required outcomes. ML pipelines can be used to construct, launch, and renew models that have already been created, allowing for a faster and more consistent transfer of findings. Model inventories and database inventories can be maintained and saved so that they can be used and reformed for other records in the future. Regular development and assessment approach and tactics may be beneficial in such a system. With auto-scaling based on GPU and CPU parameters, the system should be able to easily move accurate and securely packed models into innovation. CICD can be utilised to quickly address DevOps requirements. Natural Language Processing (NLP) may also be used in the strategy and development of backup techniques to automate the MLOps evolution [51]. The strength of the models should be tested in real-time, and maintenance should be performed to reduce the impact on the construction procedure. Furthermore, these MLOps solutions must be efficient, accessible, and dependable in practice.

In conclusion, the study shows that ML is moving away from one-on-one model verification, as described in [47], and toward more sophisticated environments in which a team of engineers works on ML research. Furthermore, even though continuous improvement is not yet a serious concern in most businesses, infrastructure issues are gaining traction. Looking at the contributor jobs, it appears that a data scientist's competencies are expanding beyond frameworks and data science to embrace additional domains, particularly ML infrastructure and use. Similar to web development, it's unclear how limited or extensive one's zone of expertise should be at this time, and the "full stack" of ML is still mostly unknown.

## IX. CONCLUSION

MLOps, or the ability to deliver ML software indefinitely, is becoming a requirement for companies that employ ML in development, comparable to DevOps for outdated software. However, because of the significant variances in functionality, DevOps technologies cannot be employed as a plug-and-play way for MLOps. Data and its challenges, as well as model development and testing, are all new challenges that necessitate new computational requirements. Furthermore, as the number of models grows, the complexity of maintaining coherence and quality grows, necessitating a well-organized versioning system for both structures and databases. To maintain object reliability, the authors stressed the need for working and effective strategies to aid with software expansion procedures employing ML methods, and DevOps. The goal of this paper is to present a review of 22 papers that have employed the MLOps concept. This study also looked at cost-effective MLOps solutions that could be used to suit the ML development phase's requirements. Although there are several MLOps platforms in use, the bulk of them has limitations when it comes to completing ML development phases and providing a mechanical structure. The evaluation of the various stages leads to a more advanced research path: the development of a fully computerised MLOps control panel for use by field professionals and decision-makers.

**Statements and Declarations**

Funding: Not Applicable.

Conflict of Interest: The authors declare no conflict of interest.

Code Availability: Not Applicable.

Availability of data and material: Not Applicable.